\documentclass{article}
\usepackage{arxiv}
\usepackage[utf8]{inputenc}
\usepackage[T1]{fontenc}  
\usepackage{hyperref}  
\usepackage{url}        
\usepackage{booktabs}    
\usepackage{amsfonts}     
\usepackage{nicefrac}      
\usepackage{microtype}      
\usepackage{lipsum}
\usepackage{graphicx}
\graphicspath{ {./images/} }
\usepackage{amsmath}

\title{Query2CAD: Generating CAD models using natural language queries}

\author{
 Akshay Badagabettu \\
  Mechanical Engineering\\
  Carnegie Mellon University\\
  Pittsburgh, PA 15213 \\
  \texttt{abadagab@andrew.cmu.edu} \\
   \And
 Sai Sravan Yarlagadda \\
  Mechanical Engineering\\
  Carnegie Mellon University\\
  Pittsburgh, PA 15213 \\
  \texttt{saisravy@andrew.cmu.edu} \\
  \And
 Amir Barati Farimani \\
  Mechanical Engineering\\
  Carnegie Mellon University\\
  Pittsburgh, PA 15213 \\
  \texttt{barati@cmu.edu} \\
}

\begin{document}
\maketitle

\begin{abstract}
Computer Aided Design (CAD) engineers typically do not achieve their best prototypes in a single attempt. Instead, they iterate and refine their designs to achieve an optimal solution through multiple revisions. This traditional approach, though effective, is time-consuming and relies heavily on the expertise of skilled engineers. To address these challenges, we introduce Query2CAD, a novel framework to generate CAD designs. The framework uses a large language model to generate executable CAD macros. Additionally, Query2CAD refines the generation of the CAD model with the help of its self-refinement loops. Query2CAD operates without supervised data or additional training, using the LLM as both a generator and a refiner. The refiner leverages feedback generated by the BLIP2 model, and to address false negatives, we have incorporated human-in-the-loop feedback into our system. Additionally, we have developed a dataset that encompasses most operations used in CAD model designing and have evaluated our framework using this dataset. Our findings reveal that when we used GPT-4 Turbo as our language model, the architecture achieved a success rate of 53.6\% on the first attempt. With subsequent refinements, the success rate increased by 23.1\%. In particular, the most significant improvement in the success rate was observed with the first iteration of the refinement. With subsequent refinements, the accuracy of the correct designs did not improve significantly. We have open-sourced our data, model, and code.\footnote{github.com/akshay140601/Query2CAD}
\end{abstract}

\section{Introduction}
\begin{figure*}
\centering
\includegraphics[width=1\textwidth]{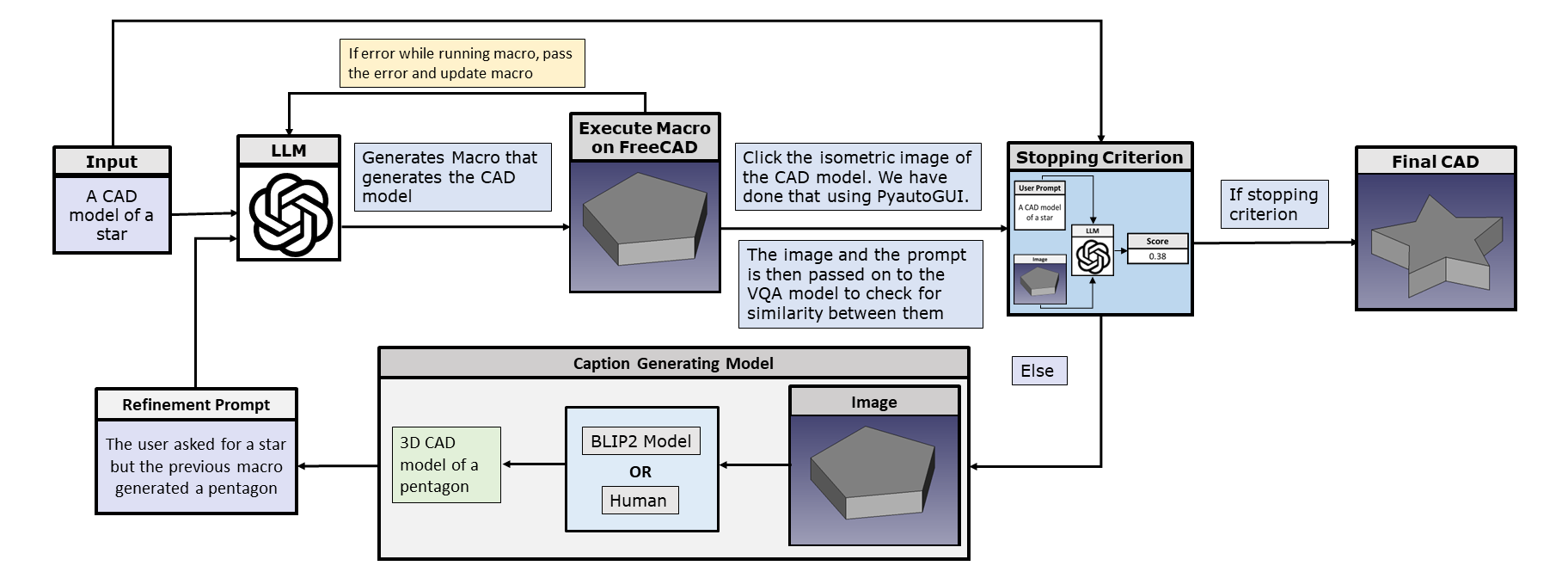}
\caption{Proposed architecture of Query2CAD. The user query is passed to an LLM that generates a Python macro to generate the corresponding CAD model. The isometric view is captured, and refinement is performed if the VQA score does not cross the threshold. The loop is run for a maximum of 3 times.}
\label{arch}
\end{figure*}

CAD has revolutionized the way engineers, architects, and designers conceptualize and design prototypes and products. CAD designing is a very important step to design prototypes and also it helps in understanding the potential failures that might occur during the manufacturing phase. Over the years, CAD softwares have evolved significantly and become very powerful. CAD designing was always done by a CAD engineer as it required a deep understanding of complex operations \cite{kasik2005ten}. The reasoning capabilities of LLMs have simplified many complex problems \cite{vaswani2017attention}. However, using LLMs to aid in the generation of CAD models remains largely untapped. The integration of natural language processing into CAD can significantly improve the design time. It can also make complex CAD software more accessible to people with less experience in CAD designing. \\\\
Over the past decade, deep learning methods had significant success in understanding data formats like mesh \cite{kalogerakis20173d}, voxels \cite{feng2019meshnet}, and point clouds.  With the advent of diffusion models, a significant shift occurred towards image-conditioned 3D generative modeling, particularly enhancing the capabilities in point clouds and neural fields \cite{melas2023pc2}. A critical limitation identified in these studies was the necessity for precise camera poses, which constrained the applicability and flexibility of the models. Recent innovations eliminated the need for camera poses \cite{xu2024instantmesh}, but the major drawback of such data format is the problems that it might pose during feature extraction. These extracted features are critical to the CAD model, and converting them to the right data format, such as triangular meshes, can be expensive and can also lead to information loss \cite{chen2018assessing}.\\\\
In this work, we introduce Query2CAD, a novel architecture that returns a CAD model similar to the user's prompt. We plan to exploit the capabilities of language models to generate Python programs (Macro) that can be executed in the FreeCAD software to obtain 3D CAD models. A designer refines his design multiple times based on the feedback given to him until he designs a CAD model that closely aligns with what the user has asked for. To mimic this behavior, our architecture has a refinement loop that refines its generation with respect to the user prompt. We incorporated a caption-generating model to provide feedback on what was designed, and, to encounter the false negatives of this model, we have also incorporated human-in-the-loop feedback. We have used Visual Question Answering Score \cite{lin2024evaluating} to check if we have reached an optimal CAD design. At a high level, our architecture takes input from the user (for instance, "Give me the CAD design of a water bottle") and passes it to a language model, which generates a working CAD design. Throughout this process, users do not require any prior knowledge of CAD software and can interact with the system using natural language. A brief overview of our architecture is shown in Figure \ref{arch}. 

\section{Related works}
\textbf{Generating 3D models:} Early explorations of generating 3D models used variational autoencoders \cite{henderson2020leveraging} \cite{henderson2020learning}. However, this just worked for simple models, and the 3D models generated were very similar to that of the training corpus and failed at generating models outside of the training corpus \cite{mi1812probe}. Later researchers have investigated 3D GANs \cite{nguyen2020blockgan} \cite{gao2022get3d}, but the main drawback of generating 3D models using GAN is its instability during training \cite{welfert2023addressing}. However, recently image conditioned 3D modeling has shown good results in generating 3D models\cite{sbrolli2022ic3d} but the precision is not very good when compared to that of a CAD model. In our research, we aim to advance beyond point cloud generation to focus on 3D CAD model creation. CAD models are pivotal to the manufacturing process and facilitate the direct transfer of data to production machinery. While 3D models generated by various image-to-3D diffusion models have accurately captured the geometry of points in the 3D CAD models \cite{edwards2024sketch2prototype}, CAD models provide more profound utility by enabling an advanced level of design, analysis, and manufacturability. 
\\\\
\textbf{Innovative approaches taken by people that can assist in 3D CAD model generation: } Recent studies, such as those by \cite{ma2023conceptual}, have highlighted the capability of language models in generating conceptual designs. These studies demonstrate that language models can effectively reason and produce detailed concept designs that can significantly assist design engineers in their work. Some works \cite{liu20233dall} integrated imagery of the model into their 3D CAD design software, enhancing the design process by inspiring more innovative ideas among designers as they create their models. However, our research aims to directly generate 3D CAD models based on prompts provided by users. 
\\\\
\textbf{Self-refinement Loops: }The iterative process used in human problem-solving, such as in CAD design, typically involves creating an initial model and refining it based on identified flaws, a method rooted in broader problem-solving strategies discussed by \cite{simon1962architecture}. Unlike this human approach, most current language models generate outputs in a single step, which contrasts with the iterative nature of human task performance \cite{madaan2024self}, \cite{baevski2018adaptive}. Although some research, such as \cite{omelianchuk2020gector}, has explored post-inference editing, these methods typically do not support multi-step refinement. Recent advancements include the development of models that integrate iterative feedback mechanisms to enhance output quality \cite{madaan2021think}. This model iteratively refines its output by reprocessing its initial output through the model to obtain and incorporate feedback internally. We aim to incorporate this approach into our architecture, thereby emulating the way CAD designers iteratively refine their prototypes.
\\\\
Human evaluation has been the gold standard in many evaluation tasks. \cite{parakh2023lifelong}, \cite{pan2023automatically} integrated human feedback during generation steps to directly address errors made by the models during generation. Inspired by such approaches, we see strong capabilities in achieving optimal generation results. Thus, we have incorporated it into every loop of our Query2CAD architecture.

\section{Preliminaries}
In this section, we have discussed more about some important components used in various modules of Query2CAD. \newline \\
\textbf{Advancements in Large Language Models for Code Generation: }Recent advances in language modeling have been propelled by the introduction of pre-trained Transformer models \cite{vaswani2017attention}, such as BERT \cite{devlin2018bert} and GPT \cite{radford2018improving}. LLMs like ChatGPT \cite{ray2023chatgpt} have demonstrated exceptional capabilities in code generation. This progress has given rise to specialized LLMs focused on software engineering, known as Code LLMs. Typically, these Code LLMs evolve from general-purpose LLMs through a process of fine-tuning, with their effectiveness being dependent on the foundational LLMs. \\\\
Notably, models such as GPT-3.5 Turbo, GPT-4 Turbo and CodeLLAMA have exhibited advanced capabilities in this area. GPT-3.5 Turbo, refined through advanced training techniques like Reinforcement Learning from Human Feedback (RLHF) \cite{ouyang2022training}, excels in producing precise and contextually relevant code outputs. This is critical in environments where integration with existing systems or adherence to specific programming standards is important. The model's ability to understand and process user instructions in natural language and convert them into functional code has made it a valuable tool in domains requiring high accuracy and detailed command execution. In benchmark tests like HumanEval, GPT-4 Turbo and CodeLLAMA have shown superior performance in code generation tasks \cite{achiam2023gpt}. By incorporating these LLMs, our system can facilitate more dynamic and interactive design processes, effectively reducing the time and expertise required to execute complex CAD design tasks.
\\\\
\textbf{Stopping criteria: }We have used Visual Question Answering Score (VQAScore) \cite{lin2024evaluating} as our stopping criteria. VQA systems have significantly advanced the integration of textual and visual data, providing a robust framework for evaluating the alignment between text descriptions and generated images. These systems utilize multimodal LLMs that effectively combine and process both modalities, enhancing the ability to accurately interpret and respond to complex queries about visual content.
The core functionality of VQA involves transforming text descriptions into specific questions, typically formatted to elicit yes-or-no answers, which are then processed alongside images. This interaction is mediated by bidirectional encoders, allowing the textual and visual inputs to dynamically influence each other's processing within the model. The resulting output is a probability distribution over potential answers, with a focus on the likelihood of "Yes," indicating a strong alignment between the image and its textual description. The VQAScore, derived from this probability, quantifies the degree of text-image congruence, offering a direct measure of the quality and relevance of the visual output in relation to the text prompt. Equation \ref{eq:vqa} gives the mathematical representation of the VQAScore.

\begin{equation}
P(\text{"Yes"} \mid \text{image},\ ''Does\ this\ figure\ show\ \{user\_query\}?\ Please\ answer\ yes\ or\ no.'')
\label{eq:vqa}
\end{equation}

\textbf{Caption Models in Computational Design:}
BLIP2 (Bootstrapped Language-Image Pre-training 2) \cite{li2023blip} is a caption model with a dual encoder-decoder architecture. This architecture effectively processes visual and textual inputs simultaneously, enabling BLIP2 to generate detailed and contextually accurate captions. BLIP2 employs transformers for handling complex data sequences and utilizes contrastive learning to align textual and visual representations closely, which improves its caption relevance and accuracy over other models.
\\
The model's superior performance is also due to its extensive pre-training on diverse datasets, allowing it to understand various visual contexts and descriptions comprehensively. BLIP2 generates captions by encoding images and texts, merging these into a joint representation, and then decoding this representation into precise captions. This capability makes BLIP2 particularly valuable in design environments, facilitating automated documentation and insightful feedback during the design process. 

\section{Methodology}

We propose Query2CAD: A novel framework to generate 3-dimensional CAD models using only natural language instructions/queries. The proposed architecture is shown in Figure \ref{arch}. The individual components of the architecture have been explained in the previous section. We now delve deeper into how our proposed system works as a whole.
\\\\
The user query is passed into a strong LLM (we have tested out our architecture using GPT-3.5 turbo and GPT-4 turbo). The LLM first generates the steps required to make the model in natural language. Using these steps it then generates a Python macro that can be executed on the FreeCAD software. FreeCAD \cite{githubFreeCAD} is an open-source CAD modeling software. If there is any error encountered when running the macro, the first type of refinement, viz., error refinement, is performed. The error message and the Python code are passed to the LLM to rectify the error and get an executable code. The error refinement is done for a user-defined $error\_iter$ number of times (default is 3).
\\\\
Once an executable code is obtained, the macro is executed on the FreeCAD software, and the isometric view of the CAD model is captured. The similarity or VQAScore between the user query and the generated isometric image is calculated by using Clip-FlanT5-XL as the VQA model. If the VQA score exceeds a user-defined threshold (default is 0.9), then the process stops, and the generated model is the final CAD design for the specified user query. But, if the threshold is not crossed, the second type of refinement, viz., model refinement, is performed. The previously captured isometric view of the CAD model is given a caption either by strong caption-generating models such as BLIP2 or by the user. The caption essentially encompasses what the LLM generated. This caption is passed as feedback to the LLM along with the generated Python code and the user query. The LLM finds the difference between the feedback and the user query and generates a modified version of its code, trying to correct its mistake. This self-refinement continues until the VQA score crosses the threshold or the user-defined number of model refinement iterations is reached (default is 3). Currently, evaluation is done manually by checking the generated 3D model and its corresponding prompt. The entire process of navigating the FreeCAD software, i.e., opening macro, executing the code, capturing screenshots, and closing the software, has been automated using PyAutoGUI \cite{githubGitHubAsweigartpyautogui}, which is a very useful Python library that can control the keyboard and mouse. Few-shot prompting was done when writing all the prompts whenever the LLM was called. The prompts can be seen in \href{https://github.com/akshay140601/Query2CAD/blob/main/src/prompts.py}{link}

\section{Experiments and Results}
\subsection{Dataset}
\label{dataset}

As there is no dataset for this particular task, we created an experimental dataset composed of questions with varying levels of difficulty, segmented into easy, medium, and hard categories. The easy section includes straightforward shapes like cubes or spheres. For instance, a typical question might be, "A CAD design of a cube with a side length of 10mm." Medium-difficulty questions involve basic shapes coupled with elementary operations, such as "A cube with a side length of 10mm positioned atop a sphere with an 8mm radius." For such inquiries, the LLM must not only generate the respective shapes but also discern their spatial arrangement. The hard category tackles more intricate designs and operations. An example of a challenging question is, "A CAD design of a basketball hoop." These questions demand that the LLMs model complex structures and comprehend the positioning of each component within the design. Our curated dataset comprises approximately 21 easy questions, 20 medium questions, and 16 hard questions. Totally, in our curated dataset, we have 57 user queries.

\subsection{Results}
\begin{figure*}[h]
\centering
\includegraphics[width=\textwidth]{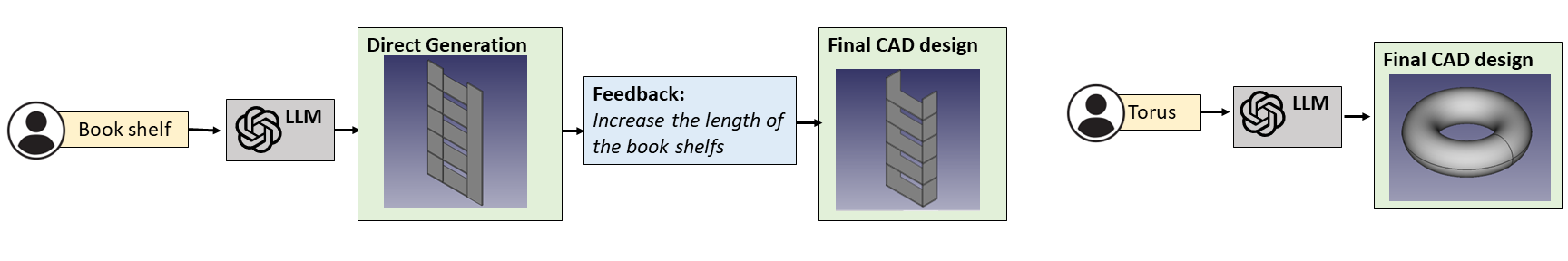}
\caption{The user query was to make a book shelf and a torus respectively. The book shelf was designed correctly within 1 refinement whereas the torus was designed correctly in the first attempt (direct generation)}
\label{fig:imageshaft}
\end{figure*}

\begin{figure*}[h]
\centering
\includegraphics[width=\textwidth]{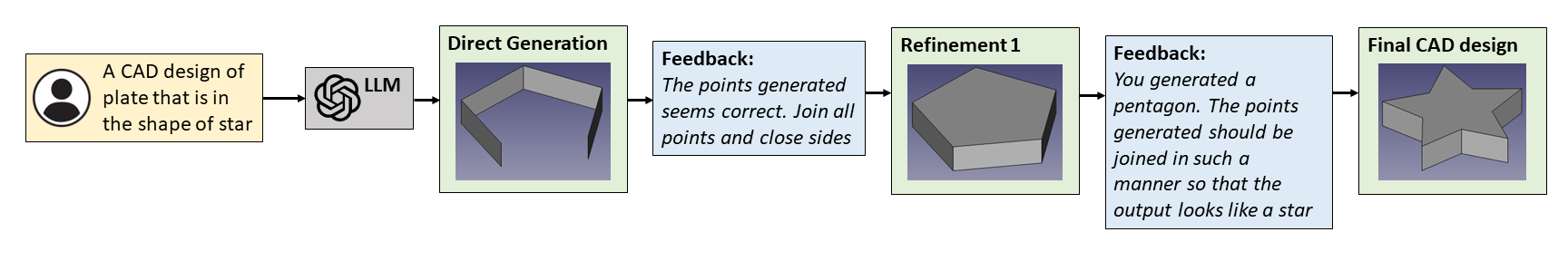}
\caption{The user query was to make a plate in the shape of a star. The LLM first generated a code that made a pentagon shape. The feedback given was to close that shape. It then made a closed pentagon with some thickness. The second feedback given was to alter the shape to a star. A star-shaped plate was then obtained}
\label{imagestar}
\end{figure*}

Query2CAD achieved great performance on easy-level questions, achieving an impressive accuracy of 95.23\% using GPT-4 Turbo as the LLM. Additionally, the model maintained commendable accuracy levels for medium and hard questions, at 70\% and 41\%, respectively. We have strategically selected GPT-4 as our language model, serving dual roles as both generator and refiner. When we used GPT-3.5-Turbo as our language model we observed that the accuracy on easy questions was 85.71\% and that of medium and hard was 35\% and 37.5\% respectively. We define an output as correct only if it produces the exact 3D CAD model as specified by the user. Any deviation from the specified design, no matter how minor, is considered a failure.
Table \ref{tab:easymedhard} displays the success rates achieved by Query2CAD when evaluated across the dataset described in Section \ref{dataset}. It also shows the success rates across various user queries with varied difficulty levels. Table \ref{tab:iter} shows the improvements shown by Query2CAD across various refinement loops. Figures \ref{imagestar} and \ref{fig:imageshaft} illustrate two outputs generated by Query2CAD, including any necessary refinements to the CAD model and the corresponding feedback provided to the model. 

\begin{table*}[h]
  \centering
  \begin{minipage}{0.48\textwidth}
    \centering
    \begin{tabular}{llll}
      \toprule
      \multicolumn{4}{c}{Difficulty level} \\
      \cmidrule(r){2-4}
      Model     & Easy     & Medium & Hard \\
      \midrule
      GPT-3.5 Turbo & 85.71\% & 35\% & 37.5\%  \\
      GPT-4 Turbo & 95.23\% & 70\% & 41.7\% \\
      \bottomrule
    \end{tabular}
    \caption{Query2CAD results on generating correct 3D CAD models using GPT-3.5 as both, base and the refiner. Metrics used for this task are defined in Section}
    \label{tab:easymedhard}
  \end{minipage}%
  \hspace{0.03\textwidth}
  \begin{minipage}{0.48\textwidth}
    \centering
    \begin{tabular}{lllll}
      \toprule
      Model & {$y_0$} & $y_1$ & $y_2$ & $y_3$\\
      \midrule
      GPT-3.5 Turbo & 32.7\% & 44.8\% & 51.7\% & 53.4\%\\
      GPT-4 Turbo & 53.6\% & 73.2\% & 76.7\% & 76.7\%\\
      \bottomrule
    \end{tabular}
    \caption{
     $y_0$ refers to the success rate of Query2CAD when the user query is passed directly to the LLM, and $y_1$, $y_2$, and $y_3$ refer to the success rate of Query2CAD after the first, second, and third refinements, respectively. 
    }
    \label{tab:iter}
  \end{minipage}
\end{table*}

\section{Analysis}
In the following section, we analyzed some results generated by the architecture in finer detail and the importance of each module in our architecture.

\begin{figure}[h]
\centering
\begin{minipage}[b]{0.45\textwidth}
\centering
\includegraphics[width=\textwidth]{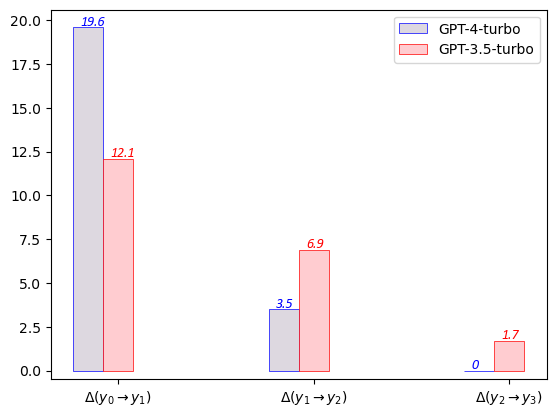}
\caption{The bar chart shows the observed improvement in success rate after every iteration. $\Delta (y_0\rightarrow y_1)$ refers to the improvement in success rate from direct generation to first iteration. Similarly $\Delta (y_1\rightarrow y_2)$ and $\Delta (y_2\rightarrow y_3)$ refers to the improvement in success rates in subsequent iterations. All the values are in percentages.}
\label{fig:iter_plot}
\end{minipage}
\hfill
\begin{minipage}[b]{0.45\textwidth}
\centering
\includegraphics[width=\textwidth]{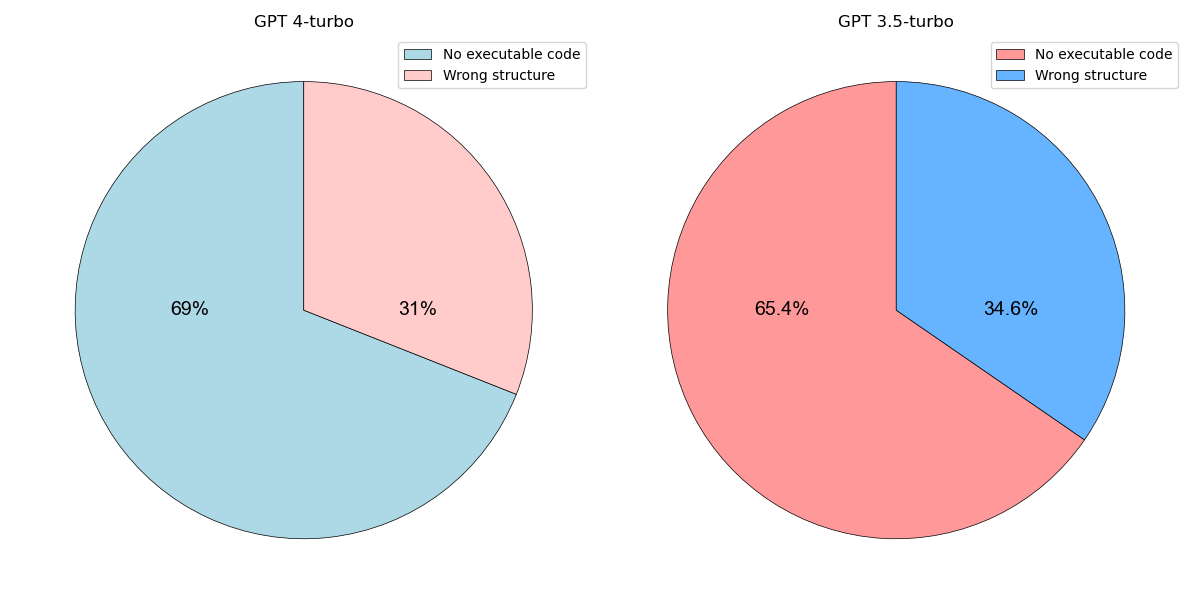}
\caption{69\% of the 13 failed cases when using GPT-4-turbo as the LLM was due to not getting an executable code and the remaining 31\% of failures were due to generating the wrong structure. Similarly, when using GPT 3.5-turbo as the LLM, 65.4\% of failures were due to not getting an executable code and 34.6\% of failures were due to generating the wrong structure}
\label{fig:failtypes}
\end{minipage}
\end{figure}

\textbf{How important are the multiple model refinement iterations?} Table \ref{tab:iter} shows the accuracy of the system for the initial output i.e., direct generation ($y_0$), and the three refinement iterations ($y_1$, $y_2$, $y_3$). We see that, on average, the accuracy improves as the number of model refinement iterations increases. But, we also notice only marginal improvements after the first refinement. This shows the importance of the first refinement, and the trend we see here was the same trend seen in \cite{madaan2024self}, albeit for different tasks. However, for a few test cases with GPT-4 as the base model, it was seen that each model refinement iteration significantly improved the 3D model, but, the number of refinement iterations was not enough to generate the exact model. Figure \ref{fig:iter_plot} shows the improvement in success rates after each iteration shown by both GPT-4 turbo and GPT-3.5 turbo. This plot clearly shows the importance of the first iteration.
\\\\
\textbf{Performance of the system on various difficulty levels of the query:} Table \ref{tab:easymedhard} shows the accuracy of the system on easy, medium, and hard levels of user queries. Refer to Section \ref{dataset} for more information on the dataset. We see that the system performs very well on the easy questions, but there is a stark decrease in performance when it comes to the medium and hard queries especially when using GPT-3.5 turbo as the base model. This trend is expected, and performance can be improved by finetuning the LLM on FreeCAD macros. When using a stronger model (GPT-4 turbo), we see a huge increase in the success rate for both the easy and medium queries.
\\\\
\textbf{Performance of the system with weaker models:} The results shown in Table \ref{tab:easymedhard} and Table \ref{tab:iter} are results of experiments performed using two of the strongest models available. Codellama-70b \cite{roziere2023code} was used as the base model to check the impact on the performance when using a weaker model. We observed extremely bad generated outputs by using a weaker model. Codellama was not able to perform reasoning very well, and sometimes, it even failed to give consistent responses to the same query. We also noticed a lot of hallucinations \cite{ji2023survey}.
\\\\
\textbf{What were the reasons behind the failed cases?} Figure \ref{fig:failtypes} shows the two kinds of failures that were observed. When using GPT-3.5 turbo as the base model, there were a total of 26 failed cases out of 57 data points, and 65.4\% of the failed cases were because of not getting an executable code even after doing error refinements. The remaining 34.6\% of the failed cases were because the model did not generate the correct structure (user query and generated output did not match even after the model refinements). The same trend can be seen when using GPT-4 turbo as the base model although there were only 13 errors. This shows that most of the errors are due to not getting an executable code and there can be some potential improvement in the way we perform error refinements.
\\\\
\textbf{Was the BLIP2 caption model effective?} During the experiments, we performed human-in-the-loop feedback wherein the BLIP2 model generates the caption first, and the user can intervene and provide their own caption if necessary. We noticed that the model was not able to refine very well with the caption generated only with BLIP2. Even though most of the time, the BLIP2 caption was perfect, we observed that providing what the generated model looks like and the steps to correct it made refinement much better. Currently, providing human feedback is much better, but it makes the system less autonomous.

\section{Limitations and future work}
One major limitation of our system is that it works very well only with strong models that are not open-source at this point of time. This is because, in our system, the LLM must be capable of both complex code generation and reasoning.\\
Currently, our curated dataset has only 57 queries. Even though we have queries of all the difficulty levels, increasing the size of the dataset can provide more insight into the actual performance of the system.\\
Human feedback works much better than any caption-generating model. The performance of the system is much better if we give the LLM what it generated and what it should do to correct itself. While this works great, the system becomes less autonomous. A promising future direction is to also input the user query in text format and either a corresponding sketch or image of the query.

\section{Conclusion}
We introduce Query2CAD, a novel approach for generating 3D CAD models. Query2CAD addresses this challenge by creating a macro that, when executed, produces a 3D CAD model. This model employs an iterative process through its architecture to ensure precise output generation. It also includes a feedback loop that employs either a BLIP2 model to describe the resemblance of the previous 3D output or direct human input to make changes to the previous design generated. We have curated a dataset comprised of diverse user queries. Upon testing these queries with Query2CAD, we found that our model excels at generating 3D models of simple shapes with an impressive accuracy rate of 95.23\%. For medium and hard-level queries, the model maintains a success rate of 70\% and 41\% respectively. Notably, our model demonstrates a significant improvement in success rate during its initial refinement phase. To this end, we make all our code, data, and prompts available at \href{https://github.com/akshay140601/Query2CAD}{link}.

\bibliographystyle{unsrt}  
\bibliography{references}
\end{document}